\documentclass{article}
\usepackage{spconf,amsmath,graphicx}
\usepackage{algorithm}
\usepackage{pifont}
\usepackage[noend]{algpseudocode}

\usepackage{threeparttable} % for "\tnote" macro and "tablenotes" environment
\usepackage{booktabs} 

\usepackage{multirow}
\usepackage{booktabs} 
\usepackage[hidelinks]{hyperref}
\title{SCORE: Self-supervised Correspondence Fine-tuning for Improved Content Representations}
\name{Amit Meghanani, Thomas Hain\thanks{This work was supported by the Centre for Doctoral Training in Speech
and Language Technologies (SLT) and their Applications funded by UK Research and Innovation [grant number EP/S023062/1]. This work was also
funded in part by LivePerson, Inc.}}
\address{Speech and Hearing Research Group \\ Department of Computer Science, The University of Sheffield, United Kingdom\\
\{ameghanani1,t.hain\}@sheffield.ac.uk}

\begin{document}
% \ninept

\maketitle
\begin{abstract}
 
There is a growing interest in cost-effective self-supervised fine-tuning (SSFT) of self-supervised learning (SSL)-based speech models to obtain task-specific representations. These task-specific representations are used for robust performance on various downstream tasks by fine-tuning on the labelled data. This work presents a cost-effective SSFT method named \textbf{S}elf-supervised \textbf{Corre}spondence (SCORE) fine-tuning to adapt the SSL speech representations for content-related tasks. The proposed method uses a correspondence training strategy, aiming to learn similar representations from perturbed speech and original speech. Commonly used
data augmentation techniques for content-related tasks (ASR) are applied to obtain perturbed speech. SCORE fine-tuned HuBERT outperforms the vanilla HuBERT on SUPERB benchmark with only a few hours of fine-tuning ($<$ 5 hrs) on a single GPU for automatic speech recognition, phoneme recognition, and query-by-example tasks, with relative improvements of 1.09\%, 3.58\%, and 12.65\%, respectively. SCORE provides competitive results with the recently proposed SSFT method SPIN, using only 1/3 of the processed speech compared to SPIN.
\end{abstract}
\section{Introduction}
\label{introduction}

Self-supervised learning (SSL) based pre-trained speech models such as HuBERT \cite{HuBERT}, WavLM \cite{WavLM} are becoming popular for their state-of-the-art performance on almost all speech applications. These models extract latent features that capture underlying factors of speech, such as acoustic-phonetic information, speaker information, semantic information, and more \cite{ssl-review}. These pre-trained representations are then fine-tuned for downstream application with labelled data. However, pre-trained SSL speech models may not be ideal for downstream tasks that do not align with the pre-trained objective (for example, handling overlapping speech \cite{WavLM}). One way to overcome this issue is to introduce a pre-training objective that relates to the downstream task, such as training with overlapping speech in WavLM \cite{WavLM}. However, this approach requires substantial amount of compute cost as the model is pre-trained from scratch. Another alternative falls within the realm of unsupervised or self-supervised fine-tuning (SSFT)\cite{SPIN}. SSFT is applied on top of pre-trained models to learn task-specific representations. Then the SSL models are fine-tuned with labelled data on the downstream tasks for robust performance. For example, ContentVec \cite{ContentVec} employs content preserving strategies (by
disentangling speakers) on top of pre-trained HuBERT model to learn content-specific representations. However, ContentVec is not very cost-effective as it requires 19 hrs on 36 GPUs on top of the pre-trained HuBERT \cite{HuBERT} model. Another recent SSFT approach for content-related downstream task is speaker-invariant clustering (SPIN) \cite{SPIN}, which requires a compute cost less than 1\% of ContentVec. SPIN employs speaker invariant clustering to improve content representations. The term SSFT was proposed in \cite{SPIN} to distinguish fine-tuning methods using only audio \cite{ContentVec,huang22b_interspeech} from supervised fine-tuning using labelled data \cite{wav2vec2}. 
\begin{figure*}[t!]
    \centering
    \includegraphics[width = 1.3\columnwidth]{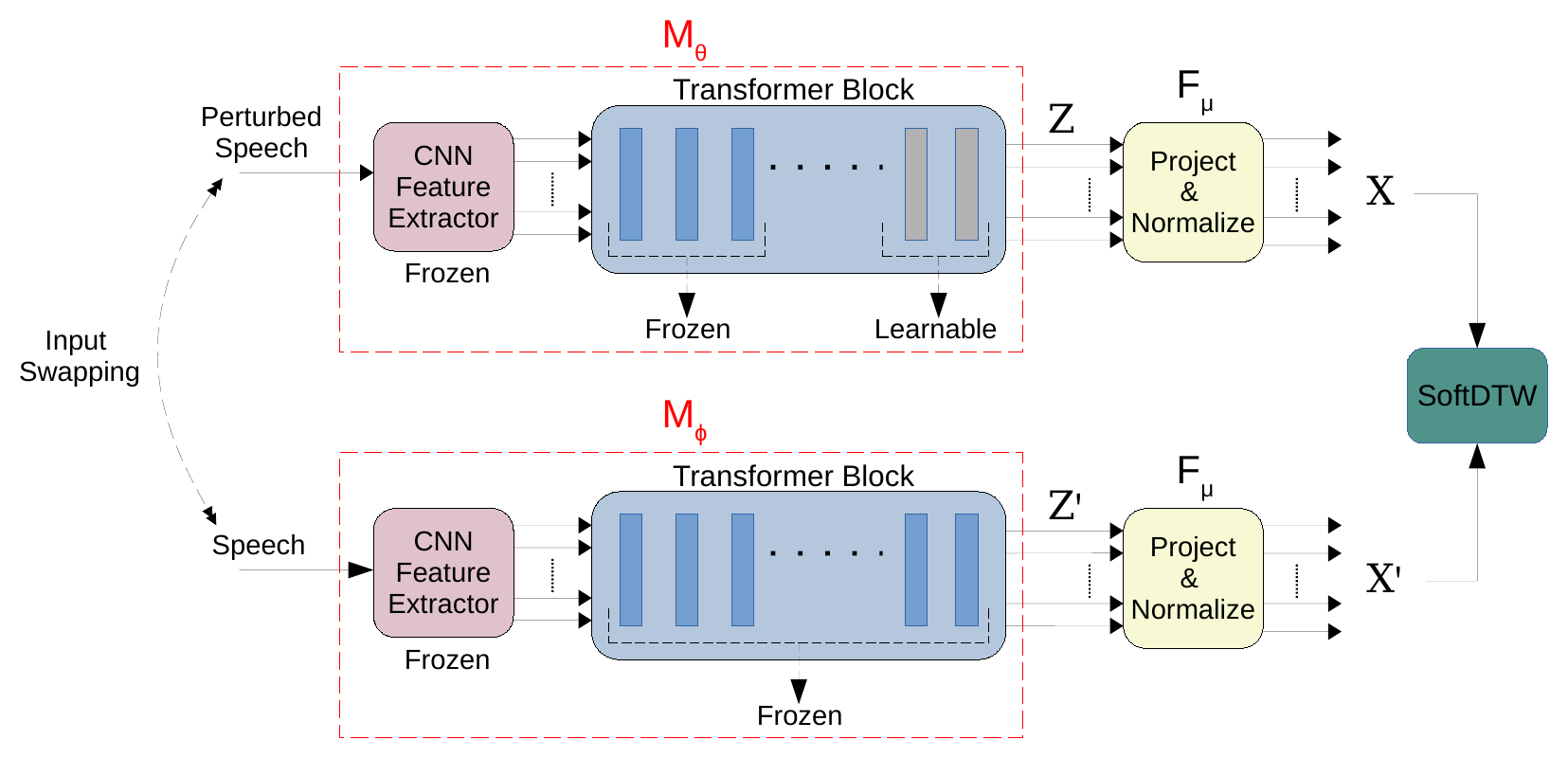}
    \caption{SCORE fine-tuning method. SCORE takes a pair of original speech and perturbed speech as input. It then matches the output sequence $X$ from the learnable model $M_\theta$ against the output sequence $X'$ from the frozen model $M_\phi$ using soft-DTW loss.}
    \label{fig1}
\end{figure*}

In this work, a simple and cost-effective SSFT method named \textbf{S}elf-supervised \textbf{Corre}spondence (SCORE) fine-tuning is proposed to preserve content. Correspondence training \cite{cae} is the task of learning similar representations from two different instances of the same spoken content. This technique has been successfully applied to extract high quality acoustic word embeddings (AWEs), where an auto-encoder takes input as a spoken word and the target output as the same word spoken by a different speaker \cite{cae,amit_asru2023}. This technique ensures that the encoder learns only content and forget other unnecessary information such as speaker, duration, prosody, etc.  Taking inspiration from this, for SCORE fine-tuning, a perturbed speech is generated from the original speech in such a way that the spoken content is preserved. Perturbed speech utterances are generated through the application of commonly employed data augmentation techniques in automatic speech recognition, such as speed perturbation \cite{data_aug_asr} and pitch shifting. After obtaining perturbed speech, the objective is to learn similar speech representations from both the original speech and perturbed speech, making the representations pitch and duration invariant. Shifting pitch (fundamental frequency) alters the speaker information while keeping the content same. To match the representations from perturbed and original speech, soft-DTW \cite{soft-DTW,soft-dtw-cuda1} is used as a loss function. Soft-DTW is a popular loss function for time-series data, and it has also been successfully used for the multi-pitch estimation task in music information retrieval \cite{mpe}. 

The proposed method is tested on three content-related downstream tasks on SUPERB benchmark \cite{superb}: automatic speech recognition (ASR), phoneme recognition (PR), and query-by-example spoken term discovery (QbE). The results are compared against the performance of vanilla models (HuBERT and WavLM) and recently proposed content-preserving SSFT methods such as ContentVec and SPIN for SUPERB benchmark . 

The main contributions of this work are as follows:
\begin{itemize}
    \item A novel cost-effective self-supervised fine-tuning method named SCORE is proposed to improve the content representations.

    \item With just less than 5 hours of SCORE fine-tuning on a single V100 GPU, SCORE fine-tuned models outperform vanilla HuBERT and WavLM on the SUPERB benchmark for content-related tasks. 

\end{itemize}
\section{Methodology}
\label{methodology}
Fig. \ref{fig1} demonstrates the proposed SCORE fine-tuning method. SCORE involves two instances of the pre-trained model, one with frozen parameters ($M_{\phi}$) and the other with learnable top layers ($M_{\theta}$), both having same initial model weights. Top layers are chosen for fine-tuning, as they encode phonetic content for most of the SSL models \cite{distill_HuBERT, layerwise_analysis}.  More implementation details are described in Sec. \ref{experiments}. The input to the SCORE is a pair of perturbed speech and original speech, randomly fed to either $M_{\theta}$ or $M_{\phi}$, as shown in Fig. \ref{fig1}. Randomizing the input ensures that the model $M_{\theta}$ does not exclusively focus on the characteristics of perturbed speech, and found to be crucial to observe the benefits of the proposed method. To obtain the perturbed speech, data augmentations used in ASR \cite{data_aug_asr} are employed, such as speech perturbation and pitch shift. Torchaudio \cite{torchaudio} is used for these perturbations, with \texttt{SpeedPerturbation} and \texttt{PitchShift} functions under \texttt{torchaudio.transforms}\footnote{\scriptsize{\url{https://pytorch.org/audio/stable/transforms.html}}}.
The obtained representations from both the models $M_{\theta}$ and $M_{\phi}$ are projected to a lower dimension with linear feedforward layers and L2-normalized. Obtained sequences from both models ($X$ and $X'$) are different in lengths due to the perturbations. Therefore, a dynamic time warping based differentiable loss function soft-Dynamic Time Warping (soft-DTW) \footnote{\scriptsize{\url{https://github.com/Maghoumi/pytorch-softdtw-cuda}}}\cite{soft-DTW,soft-dtw-cuda,softdtw_lec} is used to match sequences of unequal lengths (Eq. \ref{eq:softdtw}). This learning framework ensures that the model learns the speed and pitch invariant representations for the same spoken content. 
The soft-DTW replaces the ``min" operation in the DTW with ``soft-min" operation. Soft-DTW computes the
soft-minimum of all alignment cost. The soft-DTW for two sequences $X = x_1, x_2,...x_m$  and 
$X' = x^\prime_1 , x^\prime_2,..., x^\prime_n$ is defined \cite{softdtw_lec} as follows:

\begin{equation}
\text{soft-DTW}_{\gamma}(X, X^\prime) =
    \min_{\pi \in A(X, X^\prime)}{}^\gamma
        \sum_{(i, j) \in \pi} d(x_i, x^\prime_j)^2
\label{eq:softdtw}
\end{equation}
where $A(X, X')$ is the set of all possible paths. The $\min{}^\gamma$ is the soft-min operator with a smoothing
factor $\gamma$ and $d$ is the distance function.
The soft-min operator $\min{}^\gamma$ is defined as:

\begin{equation}
    \min{}^\gamma(a_1, \dots, a_n) = - \gamma \log \sum_i e^{-a_i / \gamma}
\end{equation}

In this work, $\text{soft-DTW}_\gamma$ is used as a loss function for SCORE fine-tuning as described in Eq. \ref{Eq1}.

\begin{equation}
\label{Eq1}
    L(X,X') = \text{soft-DTW}_\gamma(X,X')
\end{equation}

In all the experiments, we use a smoothing factor $\gamma$ of 0.1. However, to address potential negative values in $\text{soft-}DTW_\gamma$ loss, a normalized version described in Eq. \ref{Eq2} is employed. This normalization guarantees a minimum loss value of zero for identical sequences, i.e., $L_{norm}(X,X)=0$, and ensures $L_{norm}(X,X') \geq 0$ for any pair of sequences. This approach guarantees a consistently positive loss \cite{diff_divergence, tslearn}. Further, the loss from Eq. \ref{Eq2} is normalized by dividing it with  the total sequence length $m+n$. Algorithm \ref{alg:alg1} describes the entire SCORE fine-tuning method.

\vspace{-0.4cm}
\begin{equation}
\label{Eq2}
        L_{norm}(X,X') = L(X,X') - \frac{1}{2}(L(X,X) + L(X',X'))
\end{equation}

\begin{algorithm}[t!]
\small{
\caption{SCORE Fine-tuning}\label{euclid}
\begin{algorithmic}[1]
\State $M_{\theta}$ = Learnable model (Top 2 layers only)
\State $M_{\phi}$ = Frozen model
\State $M_{\theta}$, $M_{\phi}$ are initialized with the same  SSL model weights.
\State $F_{\mu}$ = Linear projection layers + L-2 Normalisation
\State $S_i$ = i\textsuperscript{th} speech utterance
\While {Not Converged}
      \For{\texttt{i=1 to $N_{samp}$}}
        \State ${S_i}^p = SpeedPerturbation(S_i)$
        \State ${S_i}^p = PitchShift({S_i}^p)$
        \State k = random(0,1)
        \If {$k == 0$}
        \State $Z = M_{\theta}({S_i}^p)$, $Z' = M_{\phi}(S_i)$ 
        \Else{}
        \State $Z = M_{\theta}(S_i)$, $Z' = M_{\phi}({S_i}^p)$ 
        \EndIf
        \State $ X = F_{\mu}(Z), X' = F_{\mu}(Z')$
        \State Compute Loss $L_{norm}(X,X')$.
        \State Compute Gradients $\frac{\partial L_{norm}}{\partial \theta}$,$\frac{\partial L_{norm}}{\partial \mu}$
        \State Update $\theta$ and $\mu$ to minimize $L_{norm}$.

      \EndFor

\EndWhile
\end{algorithmic}
\label{alg:alg1}
}
\end{algorithm}

\begin{table*}[h!]
\centering
\resizebox{0.75\textwidth}{!}{%
\begin{threeparttable}
\begin{tabular}{cccccc}
\specialrule{.1em}{0em}{0em} % Thick top rule
\multirow{2}{*}{\textbf{Model}} & \multicolumn{2}{c}{\textbf{\begin{tabular}[c]{@{}c@{}}Training\\ Processed Speech (hours)\end{tabular}}} & \multirow{2}{*}{\textbf{\begin{tabular}[c]{@{}c@{}}ASR\\ (WER) $\downarrow$\end{tabular}}} & \multirow{2}{*}{\textbf{\begin{tabular}[c]{@{}c@{}}PR\\ (PER) $\downarrow$\end{tabular}}} & \multirow{2}{*}{\textbf{\begin{tabular}[c]{@{}c@{}}QbE\\ (MTWV) $\uparrow$\end{tabular}}} \\ \cline{2-3}
                                & \textbf{Pre-training}                                   & \textbf{SSFT}                                  &                                                                               &                                                                              &                                                                                \\ \specialrule{.1em}{0em}{0em} % Thick top rule
HuBERT \cite{HuBERT}\tnote{\ding{73}}                         & 506K                                                    & 0                                              & 6.42                                                                       & 5.41                                                                         & 7.36                                                                         \\
WavLM \cite{WavLM}\tnote{\ding{73}}                          & 1439K                                                   & 0                                              & 6.21                                                                        & 4.84                                                                        & 8.70                                                                         \\ \hline
ContentVec$_{500}$ \cite{ContentVec}\tnote{\ding{73}}               & 506K                                                    & 76K                                            & 5.70                                                                          & 4.54                                                                         & 5.90                                                                         \\
HuBERT + SPIN$_{256}$ \cite{SPIN}\tnote{\ding{73}}                  & 506K                                                    & 356                                            & 6.34                                                                          & 4.39                                                                         & 9.12                                                                         \\
WavLM + SPIN$_{256}$ \cite{SPIN}\tnote{\ding{73}}                    & 1439K                                                   & 356                                            & 5.88                                                                          & 4.18                                                                         & 8.79                                                                         \\ \hline
HuBERT \cite{HuBERT}\tnote{\ding{85}}                         & 506K                                                    & 0                                              & 6.42 $\pm$ 0.08                                                                       & 5.02 $\pm$ 0.00                                                                         & 7.19                                                                         \\
WavLM \cite{WavLM}\tnote{\ding{85}}                          & 1439K                                                   & 0                                              & 6.17 $\pm$ 0.02                                                                         & 4.85 $\pm$ 0.00                                                                         & 9.15                                                                         \\ \hline
HuBERT + SCORE                   & 506K                                                    & 100                                            & 6.35 $\pm$ 0.07                                                                          & 4.84 $\pm$ 0.00                                                                      & 8.10                                                                         \\
WavLM + SCORE                    & 1439K                                                   & 100                                            & 6.15 $\pm$ 0.04                                                                          & 4.72 $\pm$ 0.00                                                                        & 9.22                                                                         \\ \hline

\end{tabular}%
    \smallskip
    \small
    \begin{tablenotes}
        \item[\ding{73}] The reported numbers are from their respective papers and SUPERB benchmark leaderboard \cite{superb} as of 13/09/2023 (\url{https://superbbenchmark.org/leaderboard}).
    \item[\ding{85}]Our results when we run the SUPERB \cite{superb} baseline scripts for HuBERT and WavLM for fair comparison.
    \end{tablenotes}
\end{threeparttable}
}
\caption{Results of the proposed SCORE fine-tuning of HuBERT and WavLM models along with baseline methods on SUPERB benchmark. The baseline methods include the BASE version of HuBERT and WavLM models, along with SSFT based ContentVec$_{500}$ and SPIN models. The downstream tasks include ASR, PR, and QbE, which are evaluated on word error rate (WER in \%), phoneme error rate (PER in \%), and maximum term weighted value (MTWV in \%), respectively. }% }. 
\label{tab1}

\end{table*}
% \vspace{-0.5cm}
\section{Experiments}
\label{experiments}
 
Experiments are conducted on two SSL speech models: HuBERT and WavLM (BASE versions). These SSL models are fine-tuned with the SCORE method. After the SCORE fine-tuning, obtained models are used for supervised training for the content-related downstream tasks on the SUPERB benchmark. 

Similar to SPIN \cite{SPIN}, the top 2 layers (11\textsuperscript{th} and 12\textsuperscript{th}) of the SSL models are fine-tuned as it is cost-effective ($ \approx 14 $M trainable parameters) and most of the SSL models encode phonetic content in top layers  \cite{distill_HuBERT,layerwise_analysis}. In this study, Wav2vec2 \cite{wav2vec2} is omitted due to the fact that the linguistic content is less well represented
in the final few layers \cite{layerwise_analysis}, which is crucial for content-related tasks. Fine-tuning the entire model, from bottom layers to top layers, would result in increased computational expenses, contradicting the study's intended objectives. Furthermore, there is a concern that when the entire model is fine-tuned, the fine-tuning objective could potentially  lead to a collapse of the original representations \cite{representation_collapse} learned during pre-training.  
The details about the data, SCORE fine-tuning, and evaluation on the SUPERB benchmark are described as following:

\noindent\textbf{Data:} In line with prior research \cite{SPIN} and to ensure a fair comparison, experiments are performed on LibriSpeech's \cite{libri} train-clean-100 hours of data for SCORE fine-tuning. Consistent with earlier discoveries \cite{SPIN},  training more layers or additional data does not enhance results.

\noindent\textbf{SCORE Fine-tuning Details:}
The representations obtained from the final Transformer layer (12\textsuperscript{th}) of the models $M_{\theta}$ and $M_{\phi}$ are sequences of 768-dimensional vectors. These vectors are projected into 256-dimensional vectors with linear projection layers and then L2-normalized. The SCORE fine-tuning method is trained for 3.6k updates ($\approx$ 1 epoch with effective batch size of 8). The model converged in just one epoch, and additional training did not yield any improvements. AdamW \cite{adamw} optimizer is used with a learning rate of $2.0e-5$ with 1k warm-up updates. One epoch roughly takes $<$ 5 hours on V100 GPU. More details are available at GitHub\footnote{\scriptsize{\url{https://github.com/Trikaldarshi/SCORE_Finetuning}}}.

 \noindent\textbf{SUPERB Benchmark:} S3PRL toolkit \footnote{\scriptsize{\url{https://github.com/s3prl/s3prl}}} is used for all the SUPERB benchmark tasks. For ASR and PR, features from all the layers are aggregated with learnable weights. These aggregated features are then fed to the prediction head for each downstream task and fine-tuned with labelled data. For ASR, the prediction head consists of 2-layer 1024-unit Bi-LSTM network with CTC loss on characters \cite{superb}. The ASR model is evaluated without any external language model. For PR, the prediction head is a frame-wise linear
transformation with CTC loss. More details can be found at SUPERB benchmark \cite{superb}. Adam optimizer is used for both ASR and PR with learning rate of $1.0e-4$ and $5.0e-4$, respectively. We conducted experiments for each ASR and PR model five times and have provided the results, including the means and standard deviations, for both the vanilla models (HuBERT and WavLM) and their SCORE fine-tuned versions. For QbE, conventional supervised phoneme posteriorgram are replaced with SSL representations \cite{superb}. For QbE, no training is required, and the evaluation is performed by running DTW on all layers separately and obtain a score for each query-document pair. For the evaluation on test set, the best layer is selected based on performance on dev set from QUESST 2014 \cite{quest2014} data. In our case, we found that 12\textsuperscript{th} layer provides best results for QbE for both HuBERT + SCORE and WavLM + SCORE. 
\label{sec:pagestyle}

\section{Results and Discussions}
\label{resultsanddiscussions}
\vspace{-0.1cm}

Table \ref{tab1} shows the processed speech during training in ``pre-training" stage and in ``SSFT stage". Processed speech is defined as ``training steps × effective batch duration" to quantify machine-independent training costs \cite{SPIN}. HuBERT + SCORE improves the HuBERT model on all three tasks with relative improvement of 1.09\%, 3.58\%, and 12.65\% for ASR, PR, and QbE, respectively. WavLM + SCORE improves the WavLM model on ASR, PR and QbE  with relative improvement of 0.32\%, 2.68\% and 0.76\%, respectively. The results are also compared with a stronger baseline ContentVec$_{500}$ \cite{ContentVec}, which uses 76K hours of processed speech compared to the SCORE which uses only 100 hrs in SSFT stage. ContentVec$_{500}$ provides better results in ASR and PR when compared with SPIN and SCORE at the compute cost of 76K hrs. 
However, both HuBERT + SCORE and WavLM + SCORE outperform ContentVec$_{500}$ on QbE task. 
Like SPIN, the goal of this work is to strike a balance between improving the downstream task and the additional training (i.e. SSFT) required. SCORE only needs $<$ 0.5 \% of processed speech when compared with ContentVec$_{500}$ in SSFT stage. SCORE provides competitive results with the SPIN models. WavLM + SCORE outperforms WavLM + SPIN$_{256}$ in QbE task. Performance of HuBERT + SCORE is close to HuBERT + SPIN$_{256}$ on ASR. 
Among all the SSFT method, SCORE uses the least amount of processed speech ($\approx$ 100 hrs) in SSFT stage.

\vspace{-0.2cm}
\subsection{Layerwise Analysis for Speaker Identification (SID)}
\vspace{-0.1cm}
One of the data augmentation techniques used in this work is pitch shift, which alters the speaker information. To assess the degradation of speaker information, experiments are conducted for SID task of SUPERB benchmark on VoxCeleb1 \cite{voxceleb1}. The same configurations are used as provided in the SUPERB \cite{superb}. Only the fine-tuned layers (i.e. 11\textsuperscript{th} and 12\textsuperscript{th}) were used for training and evaluating the SID system. The results are presented in Table \ref{tab2}. From Table \ref{tab2}, we can observe a drop in SID accuracy for both HuBERT and WavLM, in both layers.
This suggests that the SCORE fine-tuned HuBERT + SCORE and WavLM + SCORE models have representations that are relatively more speaker-invariant than the original models, benefiting content-related tasks.
\label{layerwise_analysis}

\begin{table}[h!]
\centering
\resizebox{0.70\columnwidth}{!}{%
\begin{tabular}{ccc}
\specialrule{.1em}{0em}{0em} % Thick top rule
\textbf{Model} & \textbf{Layer 11} & \textbf{Layer 12} \\ \specialrule{.1em}{0em}{0em} % 
HuBERT         & 67.73             & 64.80             \\
WavLM          & 52.30             & 49.16             \\ \hline
HuBERT + SCORE   & 66.70             & 62.61             \\
WavLM + SCORE    & 52.00             & 48.30             \\ \hline
\end{tabular}%
}
\caption{Layerwise SID accuracy (in \%) on SUPERB benchmark for original and SCORE fine-tuned SSL models.}
\label{tab2}
\end{table}

\vspace{-0.5cm}
\section{Conclusion and Future Works}
\vspace{-0.1cm}
\label{conclusion_and_future_works}
A simple and cost-effective SSFT method named SCORE is proposed to improve content representations of the pre-trained SSL speech models. For both the HuBERT and WavLM models, their respective SCORE fine-tuned models outperformed the original models on the SUPERB benchmark for ASR, PR, and QbE. Compared to other existing approaches of SSFT, SCORE requires the least amount of processed speech (less than 0.5\% of processed speech compared to ContentVec$_{500}$). SCORE provides competitive results with SPIN using 1/3 of the processed speech used by SPIN. While we observed relatively fewer improvements in ASR compared to PR and QbE, we speculate that a stronger data augmentation technique directly applicable on speech waveforms could provide better gains. We consider this research direction for our future work. 

\bibliographystyle{IEEEbib}
\ninept
\bibliography{strings,refs}

\end{document}